\documentclass[runningheads]{llncs}

\usepackage{eccv}

\usepackage{eccvabbrv}

\usepackage{graphicx}
\usepackage{booktabs}

\usepackage[accsupp]{axessibility}  
\usepackage{booktabs}
\usepackage{multirow}
\usepackage[most]{tcolorbox}
\usepackage{bbding}
\newcommand{\cmark}{\Checkmark}
\newcommand{\xmark}{\scalebox{0.8}{\XSolidBrush}}
\usepackage[table]{xcolor}
\usepackage{colortbl}
\usepackage[ruled,vlined]{algorithm2e}

\definecolor{lightgreen}{RGB}{198,239,206}
\definecolor{lightyellow}{RGB}{255,235,156}

\SetAlCapSty{textrm} 
\SetArgSty{textrm}   

\usepackage{enumitem}
\setlist[itemize]{leftmargin=*}

\usepackage[pagebackref,breaklinks,colorlinks,citecolor=eccvblue]{hyperref}

\usepackage{orcidlink}

\begin{document}

\title{Group3D: MLLM-Driven Semantic Grouping for Open-Vocabulary 3D Object Detection} 

\titlerunning{Group3D: MLLM-Based Grouping for Open-Vocabulary 3D Detection}

\author{Youbin Kim\inst{1} \and
Jinho Park\inst{1} \and
Hogun Park\inst{1} \and
Eunbyung Park\inst{2}}

\authorrunning{Kim et al.}

\institute{Department of Artificial Intelligence, Sungkyunkwan University \and
Department of Artificial Intelligence, Yonsei University}

\maketitle
\begin{figure}[tb]
  \centering
  \includegraphics[width=\linewidth]{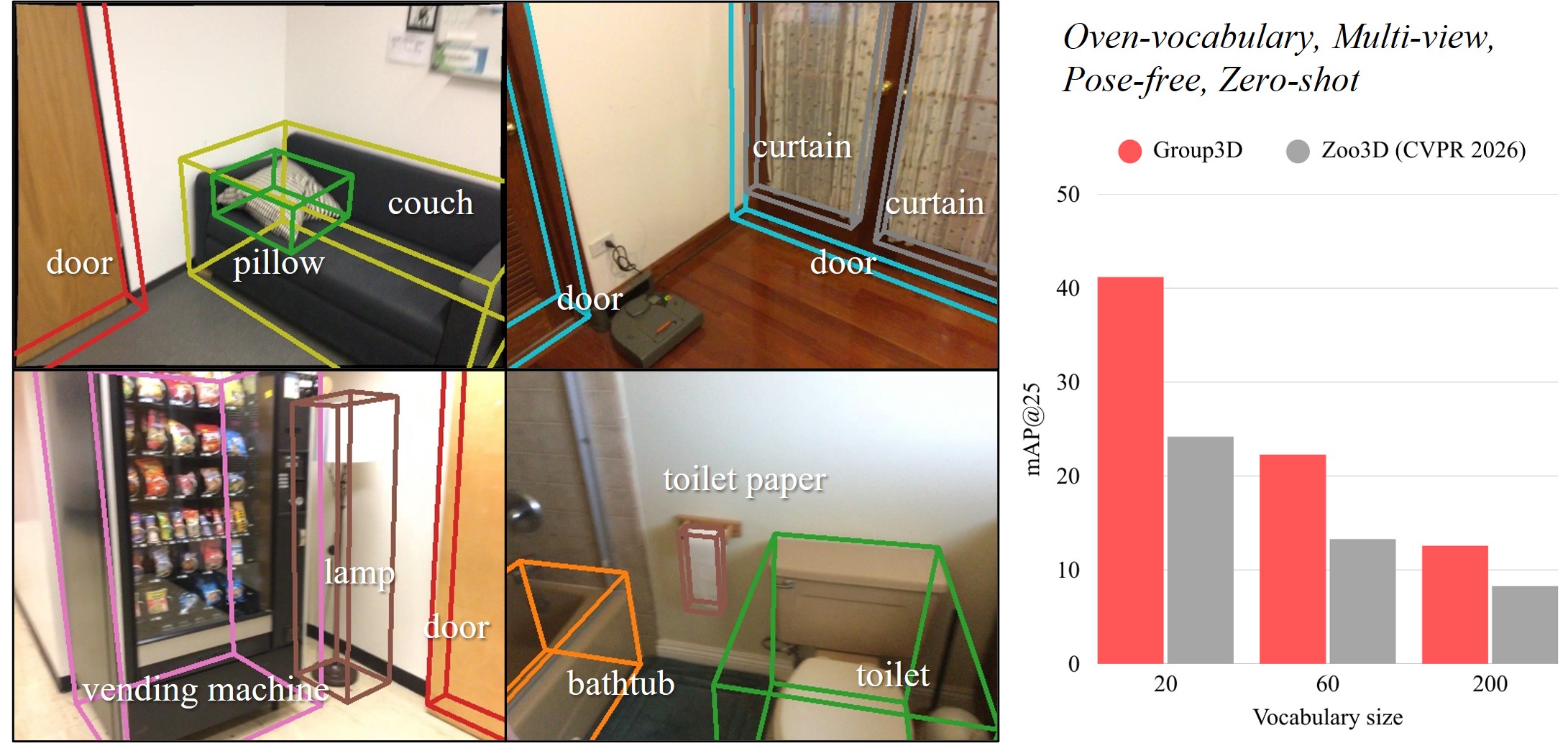}
  \caption{
\textbf{Left}: Predicted 3D bounding boxes projected onto the input RGB images.
\textbf{Right}: Comparison with the baseline under the \textit{multi-view, pose-free, zero-shot} setting across different vocabulary sizes, where Group3D consistently achieves higher mAP$_{25}$.
}
  \label{fig:example}
\end{figure}

\begin{abstract}
Open-vocabulary 3D object detection aims to localize and recognize objects beyond a fixed training taxonomy. In multi-view RGB settings, recent approaches often decouple geometry-based instance construction from semantic labeling, generating class-agnostic fragments and assigning open-vocabulary categories post hoc. While flexible, such decoupling leaves instance construction governed primarily by geometric consistency, without semantic constraints during merging. When geometric evidence is view-dependent and incomplete, this geometry-only merging can lead to irreversible association errors, including over-merging of distinct objects or fragmentation of a single instance. We propose Group3D, a multi-view open-vocabulary 3D detection framework that integrates semantic constraints directly into the instance construction process. Group3D maintains a scene-adaptive vocabulary derived from a multimodal large language model (MLLM) and organizes it into semantic compatibility groups that encode plausible cross-view category equivalence. These groups act as merge-time constraints: 3D fragments are associated only when they satisfy both semantic compatibility and geometric consistency. This semantically gated merging mitigates geometry-driven over-merging while absorbing multi-view category variability. Group3D supports both pose-known and pose-free settings, relying only on RGB observations. 
Experiments on ScanNet and ARKitScenes demonstrate that Group3D achieves state-of-the-art performance in multi-view open-vocabulary 3D detection, while exhibiting strong generalization in zero-shot scenarios. The project page is available at \url{https://ubin108.github.io/Group3D/}.

\end{abstract}

\section{Introduction}
\label{sec:intro}

3D object detection aims to localize object instances in a scene while jointly estimating their 3D position, spatial extent, and semantic identity. Beyond pixel-/point-level scene interpretation, it provides structured, object-centric representations that serve as actionable abstractions of physical environments. Such representations are a core component of modern 3D perception, enabling explicit reasoning about object geometry and spatial relationships. As language becomes increasingly intertwined with visual perception, grounding text-defined concepts to concrete 3D object instances further highlights the need for reliable instance-level 3D representations that support open-world perception.

Continuous advances in 3D geometric representation learning\cite{qi2017pointnet, zhou2018voxelnet, yan2018second, lang2019pointpillars, mao2021voxel, chen2023voxelnext, liu2025fshnet} and instance-level localization strategies\cite{qi2019deep, shi2019pointrcnn, yin2021center, liu2024seed} have substantially improved accuracy and robustness of modern 3D object detectors. Yet most existing systems are still trained within a fixed label space defined by a predefined category taxonomy and dense 3D bounding-box annotations. Consequently, detectors remain tightly coupled to the training vocabulary, and extending recognition to new object types typically requires collecting and annotating additional 3D boxes—making scale-up costly and slow.

Open-vocabulary 3D object detection mitigates this limitation by relaxing the dependence on a fixed training taxonomy and enabling recognition beyond predefined class lists. In 2D, such capability has been enabled by large-scale vision–language alignment models\cite{radford2021learning, jia2021scaling}, which learn transferable semantics from image–text data. Extending this paradigm to 3D, existing approaches often transfer open-vocabulary signals from 2D models to generate pseudo 3D supervision for training 3D detectors. Although this reduces the need for manual 3D bounding box annotations, these pipelines generally assume access to explicit 3D geometry (e.g., point clouds) for proposal generation and localization. This assumption limits applicability in scenarios where acquiring dense 3D measurements is expensive or impractical.

As an alternative, multi-view image-based 3D detection leverages inexpensive and widely available RGB observations across views. Recent multi-view open-vocabulary 3D detection pipelines often construct 3D instances in a class-agnostic manner and incorporate semantic information only after instance formation or at the representation level. While such designs simplify open-vocabulary labeling and maintain geometric robustness, they leave merging decisions governed primarily by geometric consistency. In multi-view RGB settings, geometric evidence is inherently view-dependent and often incomplete compared to ground-truth point clouds. As a result, geometry-driven merging under such ambiguity can fuse fragments that correspond to different semantic categories. Once boundaries are collapsed during instance construction, subsequent semantic reasoning may struggle to disentangle them reliably.

Building on this observation, we propose Group3D, a multi-view open vocabulary 3D object detection framework that integrates semantic and geometric cues during instance construction. Group3D operates on RGB observations of a single indoor scene and predicts a set of 3D object instances with open-vocabulary categories and 3D bounding boxes. Importantly, our approach is applicable in both pose-known and pose-free settings: when camera poses are available, Group3D directly leverages them for 3D lifting, while in the more challenging pose-free case it relies on reconstruction-based pose and depth estimates. Across both settings, the key objective is to prevent irreversible instance construction errors caused by incomplete or view-dependent geometry by enforcing semantic compatibility at merge time rather than only after instances are formed.

Group3D builds two scene-level memories to support open-vocabulary instance formation. First, it constructs a Scene Vocabulary Memory by querying a multimodal large language model (MLLM) across views, and aggregating them into a scene-adaptive vocabulary. Second, it constructs a 3D Fragment Memory by lifting category-aware 2D masks into 3D using multi-view geometry. This yields 3D fragments that preserve category hypotheses, confidence, and provenance, providing the atomic units for downstream instance construction.

Crucially, Group3D uses the MLLM to partition the scene vocabulary into semantic compatibility groups that capture plausible cross-view category variability. These groups induce a category-to-group mapping that gates fragment association. During instance formation, fragments are merged only when they satisfy both semantic compatibility and voxel-level geometric consistency. The resulting instances aggregate multi-view category evidence via confidence-weighted support statistics to select final open-vocabulary categories. As a result, Group3D achieves state-of-the-art performance in multi-view open-vocabulary 3D detection on both ScanNet~\cite{dai2017scannet} and ARKitScenes~\cite{baruch2021arkitscenes}, while exhibiting strong zero-shot generalization. In summary, our contributions are summarized as follows:
\begin{itemize}
\item We propose Group3D, a multi-view open-vocabulary 3D detection framework that constructs instances by jointly leveraging semantic compatibility and geometric consistency, mitigating irreversible over-merging under geometric ambiguity.

\item We introduce a novel MLLM-driven semantic grouping mechanism that exploits both open-vocabulary category prediction and language-induced compatibility priors to explicitly regulate 3D fragment association.

\item We achieve strong open-vocabulary and zero-shot 3D detection performance using only multi-view RGB inputs, without requiring ground-truth depth or 3D supervision.
\end{itemize}

\section{Related Works}
\subsection{3D Object Detection.}
\subsubsection{Point cloud-based detection.}
Early approaches processing point clouds to 3D object detection were confined to naively extending 2D detection paradigms into the 3D domain\cite{qi2017pointnet,qi2017pointnet++, shi2019pointrcnn, yang2019std,shi2020point}. However, due to the sparsity of 3D data, this direct extension led to severe computational waste and significant bottlenecks in both detection speed and accuracy. To overcome this limitation, VoteNet~\cite{qi2019deep} introduced a bottom-up architecture that integrated the Hough Voting into a deep learning framework. This has been established as the standard baseline for numerous closed-set 3D detectors\cite{cheng2021back, zhang2020h3dnet, wang2022rbgnet}. Several methods~\cite{ yang2018pixor, yan2018second, gwak2020generative, deng2021voxel, rukhovich2022fcaf3d, rukhovich2023tr3d} further shifted its focus toward voxel-based paradigms. These approaches discretize the continuous 3D space into voxels, allowing for the direct application of efficient 3D convolutional operations. 
\subsubsection{Multi-view image-based detection.}
Multi-view image-based 3D detection constructs object representations from multiple RGB observations of a scene. These methods broadly encompass bird’s-eye-view (BEV) projections\cite{li2024bevformer, li2023dfa3d, li2023bevdepth, huang2021bevdet}, DETR-based frameworks~\cite{carion2020end, liu2022petr, tseng2023crossdtr, wang2022detr3d}. Specifically, within the voxel-based paradigm, ImVoxelNet\cite{rukhovich2022imvoxelnet} constructs a 3D feature volume by directly lifting 2D image features into 3D voxel grids. Building upon this foundation, recent works~\cite{tu2023imgeonet, xu2023nerf, huang2025nerf, li2025go} have significantly advanced this approach. To further optimize the process, some methods~\cite{xu2024mvsdet, zhang2025boosting} explicitly predict and model the underlying scene geometry directly during the 2D-to-3D feature lifting phase. Despite these advances, most existing multi-view 3D detection frameworks operate under a closed-set setting, where detectors are trained to recognize a predefined set of object categories.

\subsection{Open-Vocabulary 3D Object Detection.}

\subsubsection{Point cloud-based detection.}
A large body of work extends conventional 3D detectors to support open-vocabulary recognition using point cloud inputs. 
Early approaches adopt CLIP-style semantic transfer by aligning proposal features with text embeddings~\cite{radford2021learning,zhu2023pointclip}.
Subsequent methods~\cite{lu2023open,cao2023coda,jiao2024unlocking,peng2024global,yang2024imov3d,wang2024ov} further improve detection by training open-vocabulary 3D detectors with pseudo supervision derived from 2D priors and cross-modal alignment.
While these approaches significantly improve open-vocabulary recognition, they typically require training on target-domain data and rely primarily on geometry-driven instance association.

\subsubsection{Multi-view image-based detection.}
Recent work has begun to extend multi-view image pipelines to open-vocabulary 3D detection. In these approaches, 2D predictions are lifted into 3D and aggregated across views to form object hypotheses.
OpenM3D~\cite{hsu2025openm3d} proposes an open-vocabulary multi-view detection framework trained with pseudo 3D boxes and CLIP-based semantic alignment without requiring human annotations.
Zoo3D~\cite{lemeshko2025zoo3d}, in contrast, constructs 3D boxes by clustering lifted 2D masks and assigns semantic labels via vision-language similarity.
However, these pipelines largely rely on geometric consistency for cross-view instance construction and incorporate semantic cues only after instances are formed.
Geometry-first aggregation can lead to over-merging when observations are incomplete or geometrically ambiguous.
Our method instead integrates semantic constraints directly into the instance construction process via MLLM-driven compatibility grouping, enabling more robust cross-view association.

\section{Group3D}
\label{sec:method}
\subsubsection{Problem Setup}
\label{subsec:setup}

We address multi-view open-vocabulary 3D object detection from RGB observations.
Given a set of RGB images $\mathcal{I}=\{I_n\}$ captured from a single scene, along with optional camera poses $\{{T}_n\}$,
our goal is to predict a set of 3D object instances $\mathcal{O}=\{ \big(\ell_k, s_k, {b}_k\big) \}_k$,
where $\ell_k$ denotes the predicted open-vocabulary category, 
$s_k$ is its confidence score, 
and ${b}_k$ is an axis-aligned 3D bounding box.

\begin{figure}[tb]
  \centering
  \includegraphics[width=\linewidth]{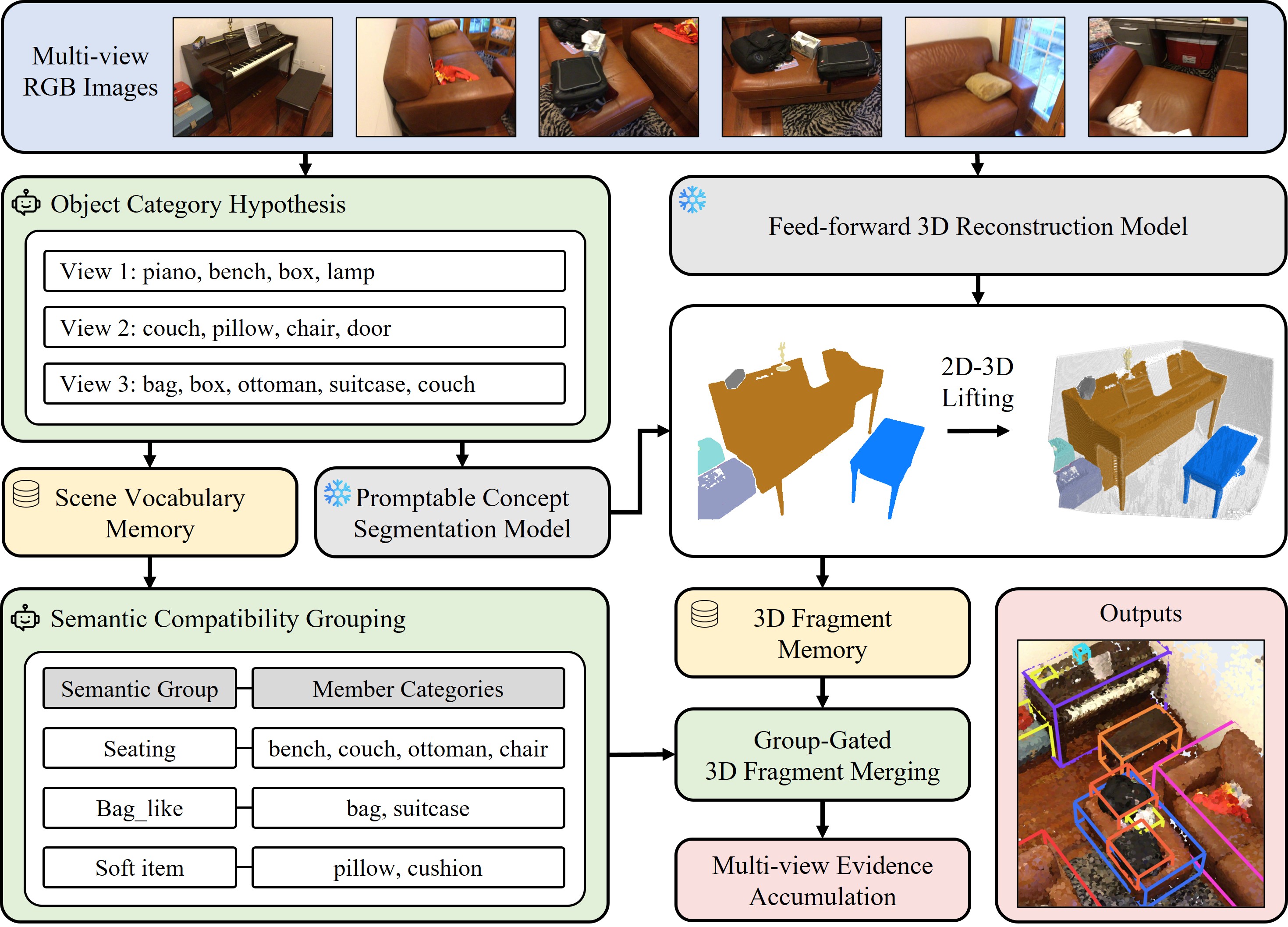}
  \caption{
The overview of Group3D.
Given multi-view RGB images, an MLLM predicts object categories across views,
which are aggregated into a \emph{Scene Vocabulary Memory}.
Category-aware masks are lifted into 3D to construct a \emph{3D Fragment Memory}.
The MLLM then organizes the vocabulary into semantic compatibility groups,
which gate fragment merging together with geometric consistency to produce
the final open-vocabulary 3D object instances. Finally, multi-view evidence is accumulated to determine the final open-vocabulary category and 3D bounding box for each object instance.
}
  \label{fig:fig_main}
\end{figure}

\subsection{Scene Memory Construction}
Group3D constructs two scene-level memories: 
(i) \emph{Scene Vocabulary Memory}, which aggregates object category hypotheses predicted across views into a compact scene-adaptive category set, and 
(ii) \emph{3D Fragment Memory}, which stores all 3D fragments obtained by lifting category-aware 2D masks into the reconstructed 3D space.

\subsubsection{Scene Vocabulary Memory.}
\label{subsec:memory}
Given an input view $I_n$, we query an MLLM to obtain a set of object categories, $\mathcal{V}_n$. 
The predicted categories are normalized through canonicalization, including casing normalization and morphological standardization, e.g., \emph{Trash\_can} $\rightarrow$ \emph{trash can}.
We then aggregate the normalized categories across views and remove duplicates to form a scene-level vocabulary, $\mathcal{V} =\bigcup_{n}{\mathcal{V}}_n$, referred to as the \emph{Scene Vocabulary Memory}, which is subsequently used to induce semantic compatibility groups (\cref{subsec:grouping}).

\subsubsection{3D Fragment Memory.}
\label{subsec:fragment}
We leverage a foundational segmentation model, SAM 3\cite{carion2025sam} to obtain category-aware 2D masks. By querying each category $\ell_{i} \in \mathcal{V}$ in the scene vocabulary, we produce 2D masks $m_{n,i} \in \{0,1 \}^{H\times W}$ for each input image $I_n$ and each category $\ell_i$, along with the confidence score $s_{n,i}$. Then, to lift 2D masks into 3D space, we obtain camera poses and depth maps using a reconstruction model applied to the input images. 
When ground-truth camera poses are available, we use them instead of the predicted poses. 
The resulting poses $\{{T}_n\}$ and depth maps $\{D_n\}$ define a shared world coordinate system for projecting 2D masks into 3D.

Each mask $m_{n,i}$ is lifted into 3D by back-projecting its pixel coordinates $\{ (u,v) \mid m_{n,i}(u,v)=1 \}$ using the obtained depth and pose, where $(\cdot,\cdot)$ denotes an indexing operator.
Let ${K}_n$ denote the camera intrinsic matrix and ${T}_n=[{R}_n \mid {t}_n]$ the camera pose mapping world coordinates to the camera frame.
We lift each mask $m_{n,i}$ into 3D via back-projection,
\begin{equation}
{p}(u,v)={R}_n^{\top}\big(D_n(u,v)\,{K}_n^{-1}
\begin{bmatrix}
u \\ v \\ 1
\end{bmatrix} - {t}_n\big),
\end{equation}
and define the corresponding point clouds fragment $F_{n,i}$, and \emph{3D Fragment Memory} $\mathcal{F}$ can be defined as follows,
\begin{equation}
\mathcal{F} := \{(F_{n,i},{\ell}_{i},{s}_{n,i})\}_{n,i}, \quad F_{n,i}=\{{p}(u,v)\in\mathbb{R}^{3} \mid m_{n,i}(u,v)=1, \forall u,v\}.
\end{equation}
To mitigate reconstruction noise, we apply reliability filtering and suppress extreme depth outliers within each mask region, and each fragment stores its 3D point clouds $F_{n,i}$, category hypothesis ${\ell}_{i}$, and confidence score ${s}_{n,i}$.

Regarding the confidence score, we defined it as the product of a query-level score and a global presence score:
\begin{equation}
{s}_{n,i} = {s}^{\text{query}}_{n,i}\cdot {s}^{\text{pres}}_{n},
\label{eq:sam3_score}
\end{equation}
where ${s}^{\text{pres}}_{n}$ estimates whether the prompted category is present in $I_n$, and ${s}^{\text{query}}_{n,i}$ measures the match of the corresponding between the prompted category and the mask region.

\subsection{Semantic Compatibility Grouping}
\label{subsec:grouping}

Open-vocabulary predictions across views can be inconsistent due to taxonomy noise, where the same physical object may receive different but semantically related categories across frames.
To transform this variability into a structured prior for instance construction, we query the MLLM to partition the scene vocabulary $\mathcal{V}$ into semantic compatibility groups,
\begin{equation}
\mathcal{G}=\{G_g\}_{g=1}^{G}, \quad G_g\subseteq\mathcal{V}.
\end{equation}
The MLLM is prompted to group categories that could plausibly refer to the same physical object under taxonomy noise (e.g., \emph{chair}–\emph{sofa}, \emph{desk}–\emph{table}), while avoiding merges that are structurally inconsistent.
In particular, categories corresponding to structural attachments (e.g., \emph{wall}–\emph{window} or \emph{wall}–\emph{door}), supporting structures (e.g., \emph{floor}–\emph{wall}), or part–whole relationships (e.g., \emph{table}–\emph{cup}) are explicitly excluded from the same group.

As a result, the induced grouping captures semantic substitutability rather than spatial adjacency or co-occurrence. These groups define which category labels are considered compatible and therefore allowed to merge across views. Candidate merges are subsequently verified using geometric consistency during instance merging.

\subsection{Group-Gated 3D Fragment Merging}
\label{subsec:merge}




\begin{algorithm}[t]
\caption{Group-Gated 3D Fragment Merging}
\label{alg:group_gated_merge}
\KwIn{\emph{3D Fragments Memory} $\mathcal{F}$}
\KwOut{Merged clusters $\mathcal{C}$ (3D instances)}

Sort fragments by descending spatial extent\;
$\mathcal{C}\leftarrow\emptyset$\;

\ForEach{$(F_{n,i},\ell_i,{s}_{n,i}) \in \mathcal{F}$}{
  merged $\leftarrow$ False\;
  \ForEach{$({C}_F, {C}_{\ell}) \in \mathcal{C}$}{
    \If{$g(\ell_i)=g({C}_{\ell}) \ \land\ \textrm{\texttt{Overlap}}(F_{n,i},{C}_F) $}{
    ${C}_F \leftarrow {C}_F \cup \{ F_{n,i} \}$\tcp*[r]{Merge point clouds fragments}
    ${C}_{\ell} \leftarrow {C}_{\ell} \cup \{ \ell_i \}$\tcp*[r]{Keep a set of open-vocab categories}
    \textrm{merged} $\leftarrow$ True\; \textbf{break}\;
    }
  }
  \If{\textrm{merged} = False}{
    $\mathcal{C}\leftarrow \mathcal{C}\cup\{(F_{n,i}, \{ \ell_i \})\}$\tcp*[r]{Add a new fragment}
  }
}
\Return $\mathcal{C}$\;
\end{algorithm}

We construct global 3D instances by merging fragments in $\mathcal{F}$ under a semantic compatibility constraint combined with geometric consistency.
The defining characteristic of Group3D is that fragment association is explicitly gated by semantic compatibility groups introduced in \cref{subsec:grouping}, rather than relying on geometry alone.
Let $g(\cdot)$ denote the mapping a category (or a set of categories) to semantic compatibility group, then two point cloud fragments $F_{n,i}$ and $F_{m,j}$ are merge-eligible only if they satisfy the following condition $g({\ell}_{i}) = g({\ell}_{j})$, i.e., both categories are in the same group, ensuring that only semantically compatible fragments can be associated.

Geometric consistency is then verified using voxel overlap.
Each fragment is represented by its voxel set $\texttt{vox}(\cdot)$, and overlap is measured using Intersection over Union (IoU) together with a containment ratio:
\begin{equation}
\text{IoU}_\texttt{vox}(A,B)=\frac{|\texttt{vox}(A)\cap \texttt{vox}(B)|}{|\texttt{vox}(A)\cup \texttt{vox}(B)|}, 
\qquad
\text{Cont}_\texttt{vox}(B\!\rightarrow\!A)=\frac{|\texttt{vox}(A)\cap \texttt{vox}(B)|}{|\texttt{vox}(B)|}.
\end{equation}
IoU alone may underestimate geometric agreement when fragments differ substantially in spatial extent.
If fragment $B$ is significantly smaller than fragment $A$, IoU can remain low even when $B$ overlaps heavily with $A$ or is almost entirely contained within it.
In such cases, the union term is dominated by the larger fragment, diluting the overlap score.
The containment ratio explicitly measures how much of the smaller fragment is supported by the larger one, thereby capturing this asymmetric inclusion.
We define $\texttt{Overlap}(A,B)$ as a boolean predicate based on these measures as follows,
\begin{equation}
\text{Overlap}(A,B) =
(\text{IoU}_{\text{vox}}(A,B) \ge \tau_{\text{iou}})
\;\lor\;
(\text{Cont}_{\text{vox}}(B\!\rightarrow\!A) \ge \tau_{\text{cont}}).
\end{equation}
Combining these conditions, the fragment merging is performed under the conjunction of semantic compatibility, geometric overlap, and cross-view consistency.
\cref{alg:group_gated_merge} summarizes the resulting group-gated merging procedure, which produces final 3D instance clusters $\mathcal{C}$.

\subsection{Multi-view Evidence Accumulation}
\label{subsec:evidence}


After group-gated merging (\cref{alg:group_gated_merge}), each 3D instance $(C_F, C_{\ell})$ contains the merged point cloud fragments $C_F$ and the set of associated category labels $C_{\ell}$. To determine the final label, we aggregate the candidate categories together with their confidence scores.
Slightly abusing notation, let $\ell \in C_{\ell}$ denote a category label associated with the 3D instance. We compute its mean confidence score $\bar{s}(\ell)$ by averaging the confidence scores of all fragments associated with the category $\ell$. Since each 3D instance is formed by merging fragments originating from multiple input views, the same category may be associated with multiple fragments, each carrying a different confidence score.
The instance-level category score is then defined as,
\begin{equation}
s(\ell)=\bar{s}(\ell)\cdot w\!\left( N(\ell) \right),
\end{equation}
where $N(\ell)$ denotes the number of fragments associated with category $\ell$ during the merging stage, and $w(x)=1-\exp\!\left(-\frac{x}{\tau}\right)$ is a monotonically increasing function that rewards repeated cross-view support while preventing disproportionate dominance by categories with many fragments.
The final instance label is selected as $\arg\max_{\ell \in C_{\ell}} s(\ell)$, with the corresponding score $s(\ell)$.
The 3D bounding box is computed by taking the minimum and maximum coordinates of $C_F$ along each axis.

\section{Experiments}
\subsection{Datasets}
\label{subsec:datasets}
Evaluation is conducted on two multi-view indoor 3D perception benchmarks, ScanNetV2~\cite{dai2017scannet} and ARKitScenes~\cite{baruch2021arkitscenes}, and results are reported on the official validation splits. Since the proposed pipeline is training-free with respect to 3D supervision, these benchmarks are used solely for evaluation. Following standard 3D object detection protocols, mean average precision (mAP) is reported at 3D IoU thresholds of 0.25 and 0.50.

\subsubsection{ScanNet.}
\label{subsec:datasets_scannet}
ScanNetV2~\cite{dai2017scannet} is a standard indoor RGB-D benchmark that provides reconstructed scenes with multi-view RGB sequences, aligned camera trajectories, and 3D instance annotations. The official split contains 1,201 training scenes and 312 validation scenes. To characterize open-vocabulary generalization across vocabulary scales, three established settings are considered, denoted as ScanNet20, ScanNet60, and ScanNet200: (i) a 20-category setting following~\cite{lu2023open}; (ii) a 60-category setting following~\cite{cao2023coda, wang2024ov}, where categories are defined by training frequency, treating the top-10 most frequent categories as seen and 50 additional categories as novel; and (iii) a 200-category setting following~\cite{rozenberszki2022language}, which expands the label space to 200 fine-grained categories with a pronounced long-tail distribution. The ScanNet60 setting is commonly used with supervised training on the seen categories; comparisons therefore include both supervised methods trained on the seen set and zero-shot methods that use no category-specific 3D supervision.

\subsubsection{ARKitScenes.}
\label{subsec:datasets_arkitscenes}
ARKitScenes~\cite{baruch2021arkitscenes} provides real-world indoor multi-view RGB-D sequences with reconstructed scene geometry and 3D object annotations for 17 object categories. The official split contains 4,493 training scans and 549 validation scans.

\begin{table}[t]
\caption{Quantitative results on the ScanNet benchmark under two category settings.
Methods are grouped by input modality, including point cloud-based methods and multi-view image-based methods.
For multi-view methods, results are further reported with and without ground-truth camera poses.
$^\dagger$ denotes methods that use 3D bounding boxes during training.
Zoo3D$_0$ and Zoo3D$_1$ denote the zero-shot and self-supervised variants of Zoo3D, respectively.}
\label{tab:scannet_improved}
\centering
\scriptsize
\setlength{\tabcolsep}{4pt}
\renewcommand{\arraystretch}{1.05}

\begin{tabular}{p{3.2cm} c c cc cc}
\toprule
\textbf{Method} & \textbf{Pose-free} & \textbf{Zero-shot} 
& \multicolumn{2}{c}{\textbf{ScanNet20}} 
& \multicolumn{2}{c}{\textbf{ScanNet60}} \\
\cmidrule(lr){4-5} \cmidrule(lr){6-7}
 &  &  
 & \textbf{mAP$_{25}$} & \textbf{mAP$_{50}$}
 & \textbf{mAP$_{25}$} & \textbf{mAP$_{50}$} \\
\midrule

\multicolumn{7}{l}{\textit{\textbf{Point cloud-based}}} \\
\midrule
Det-PointCLIPv2$^\dagger$\cite{zhu2023pointclip} & --  & \xmark & -- & -- & 0.2 & -- \\
3D-CLIP$^\dagger$\cite{radford2021learning}         & --  & \xmark & -- & -- & 4.0 & -- \\
OV-3DET\cite{lu2023open}         & --  & \xmark & 18.0 & -- & -- & -- \\
CoDa$^\dagger$\cite{cao2023coda}            & -- & \xmark & 19.3 & -- & 9.0 & -- \\
INHA$^\dagger$\cite{jiao2024unlocking}            & --  & \xmark & -- & -- & 10.7 & -- \\
GLIS\cite{peng2024global}            & --  & \xmark & 20.8 & -- & -- & -- \\
ImOV3D\cite{yang2024imov3d}          & --  & \xmark & 21.5 & -- & -- & -- \\
OV-Uni3DETR$^\dagger$\cite{wang2024ov}     & -- & \xmark & 25.3 & -- & 19.4 & -- \\
Zoo3D$_{0}$\cite{lemeshko2025zoo3d}               & -- & \cmark & 34.7 & 23.9 & 27.1 & 18.7 \\
Zoo3D$_{1}$\cite{lemeshko2025zoo3d}               & --  & \xmark & \textbf{37.2} & \textbf{26.3} & \textbf{32.0} & \textbf{20.8} \\
\midrule

\multicolumn{7}{l}{\textit{\textbf{Multi-view image-based}}} \\
\midrule
OV-Uni3DETR$^\dagger$\cite{wang2024ov} & \xmark & \xmark & -- & -- & 11.2 & -- \\
OpenM3D\cite{hsu2025openm3d}               & \xmark & \xmark & 19.8 & 7.3 & -- & -- \\
Zoo3D$_{0}$\cite{lemeshko2025zoo3d}           & \xmark & \cmark & 30.5 & 17.3 & 22.0 & 10.4 \\
Zoo3D$_{1}$\cite{lemeshko2025zoo3d}           & \xmark & \xmark & 32.8 & 15.5 & 23.9 & 10.8 \\
\textbf{Group3D} (Ours)         & \xmark      & \cmark & \textbf{51.1} & \textbf{27.4} & \textbf{29.1} & \textbf{13.9} \\

\midrule
Zoo3D$_{0}$\cite{lemeshko2025zoo3d}         & \cmark & \cmark & 24.2 & 8.8 & 13.3 & 4.1 \\
Zoo3D$_{1}$\cite{lemeshko2025zoo3d}         & \cmark & \xmark & 27.9 & 10.4 & 15.3 & 5.6 \\
\textbf{Group3D} (Ours) & \cmark      & \cmark & \textbf{41.2} & \textbf{18.5} & \textbf{22.3} & \textbf{8.5} \\
\bottomrule
\end{tabular}

\vspace{-2mm}
\end{table}

\subsection{Implementation Details}
\label{subsec:impl}
\subsubsection{Experimental settings.}
For each scene, we uniformly sample 128 frames and resize all frames to $378 \times 504$ for reconstruction, following the input resolution setting of the reconstruction backbone. We extract the $K$ category hypotheses per view for scene vocabulary construction and set $K=5$ in all experiments. We use GPT-5.1 as the MLLM for category proposal and semantic grouping. During group-gated fragment merging, we voxelize fragments with a fixed voxel size of $5cm$ to compute voxel overlap and containment. All experiments are conducted on a single NVIDIA A6000 GPU. Additional implementation details are provided in the supplementary material.


\begin{table*}[t]
\caption{Per-class AP$_{25}$ comparison on ScanNet20. `pc' indicates the methods leverage the ground-truth point clouds, `pi' denotes the use of posed images, and `ui' means the use of unposed images.}
\centering
\small
\setlength{\tabcolsep}{4pt}
\resizebox{\textwidth}{!}{%
\begin{tabular}{l|c|cccccccccc|c}
\toprule
& Inputs & toilet & bed & chair & sofa & dresser & table & cabinet & bookshelf & pillow & sink & \\
\midrule
OV-3DET 
& pc + pi & 57.3 & 42.3 & 27.1 & 31.5 & 8.2 & 14.2 & 3.0 & 5.6 & 23.0 & 31.6 \\
CoDA 
& pc & 68.1 & 44.1 & 28.7 & 44.6 & 3.4 & 20.2 & 5.3 & 0.1 & 28.0 & 45.3 \\
OV-Uni3DETR 
& pc & 86.1 & 50.5 & 28.1 & 31.5 & 18.2 & 24.0 & 6.6 & 12.2 & 29.6 & \underline{54.6} \\
Zoo3D$_{1}$ 
& pc + pi & 78.4 & 54.4 & \textbf{74.4} & 65.5 & 33.6 & 19.1 & 14.1 & \underline{32.3} & \textbf{46.1} & 27.3 \\

\rowcolor{gray!15}
\textbf{Group3D} 
& pi & {\textbf{91.3}} & {\textbf{80.5}} & {\underline{61.1}} & {\textbf{78.0}} & {\textbf{55.4}} & {\textbf{56.2}} & {\textbf{24.7}} & {\textbf{41.3}} & {\underline{43.7}} & {\textbf{62.3}} & \\
\rowcolor{gray!15}
(Ours) 
& ui & {\underline{89.6}} & {\underline{80.2}} & 37.8 & {\underline{73.4}} & {\underline{50.5}} & {\underline{43.1}} & \underline{16.6} & 31.7 & 20.2 & 39.3 & \\
\midrule
& & bathtub & refrigerator & desk & nightstand & counter & door & curtain & box & lamp & bag & Mean\\
\midrule

OV-3DET 
& & 56.3 & 11.0 & 19.7 & 0.8 & 0.3 & 9.6 & 10.5 & 3.8 & 2.1 & 2.7 & 18.0 \\
CoDA 
& & 50.5 & 6.6 & 12.4 & 15.2 & 0.7 & 8.0 & 0.0 & 2.9 & 0.5 & 2.0 & 19.3 \\
OV-Uni3DETR 
& & 63.7 & 14.4 & 30.5 & 2.9 & \underline{1.0} & 1.0 & 19.9 & 12.7 & 5.6 & 13.5 & 25.3\\
Zoo3D$_{1}$
& & 64.6 & \underline{57.5} & 10.7 & \underline{58.8} & 0.2 & \underline{27.4} & 8.0 & \underline{20.0} & \textbf{43.3} & 9.1 & 37.2\\

\rowcolor{gray!15}
\textbf{Group3D} 
& & {\textbf{85.8}} & {\textbf{63.3}} & {\textbf{65.1}} & {\textbf{70.4}} & \textbf{2.0} & {\textbf{40.8}} & {\textbf{24.1}} & {\textbf{22.6}} & {\underline{35.7}} & {\textbf{18.3}} & {\textbf{51.1}} \\
\rowcolor{gray!15}
(Ours)
& & {\underline{82.4}} & 52.2 & {\underline{60.1}} & {57.8} & {0.9} & 24.0 & {\underline{20.9}} & 15.8 & 13.7 & \underline{14.4} & \underline{41.2} \\
\bottomrule
\end{tabular}}
\label{tab:ap25}
\end{table*}

\begin{table}[h]
\caption{Quantitative results on ScanNet200\cite{rozenberszki2022language} and ARKitScenes\cite{baruch2021arkitscenes} under the multi-view setting.}
\label{tab:scannet200_arkitscenes}
\centering
\scriptsize
\setlength{\tabcolsep}{4pt}
\renewcommand{\arraystretch}{1.05}

\begin{tabular}{l c c cc cc}
\toprule
\multirow{2}{*}{\textbf{Method}} 
& \multirow{2}{*}{\textbf{Pose-free}}
& \multirow{2}{*}{\textbf{Zero-shot}} 
& \multicolumn{2}{c}{\textbf{ScanNet200}} 
& \multicolumn{2}{c}{\textbf{ARKitScenes}} \\
\cmidrule(lr){4-5}\cmidrule(lr){6-7}
& & 
& \textbf{mAP$_{25}$} & \textbf{mAP$_{50}$}
& \textbf{mAP$_{25}$} & \textbf{mAP$_{50}$} \\
\midrule
OpenM3D\cite{hsu2025openm3d}      & \xmark & \xmark & 4.2  & -- & --   & -- \\
Zoo3D$_0$\cite{lemeshko2025zoo3d}    & \xmark & \cmark & 14.3 & 6.2 & --   & -- \\
Zoo3D$_1$\cite{lemeshko2025zoo3d}    & \xmark & \xmark & 16.5 & 6.3 & --   & -- \\
Ours         & \xmark & \cmark & \textbf{17.9} & \textbf{8.7} & \textbf{20.5} & \textbf{5.9} \\
\midrule
Zoo3D$_0$\cite{lemeshko2025zoo3d}    & \cmark & \cmark & 8.3  & 2.9 & 13.0 & 2.6 \\
Zoo3D$_1$\cite{lemeshko2025zoo3d}    & \cmark & \xmark & 10.7 & 3.8 & 16.1 & 3.5 \\
Ours         & \cmark & \cmark & \textbf{12.6} & \textbf{5.7} & \textbf{18.4} & \textbf{4.5} \\
\bottomrule
\end{tabular}

\end{table}

\subsubsection{Zero-shot setting.}
All results are obtained in a zero-shot manner without using category-specific 3D supervision from the evaluated benchmarks. To avoid dataset-specific training leakage, 3D reconstruction backbones are selected such that they are not trained on the target benchmark. Accordingly, Depth Anything~3~\cite{lin2025depth} is used for ScanNetV2, and VGGT~\cite{wang2025vggt} is used for ARKitScenes. This ensures that the proposed pipeline does not rely on dataset-specific supervision from the evaluation benchmarks.

\subsubsection{Reconstruction-based geometry and alignment.}
Both pose-known and pose-free settings rely on RGB-only reconstruction for 3D lifting; in the pose-known setting, the provided camera poses are used in place of estimated poses. Following Zoo3D~\cite{lemeshko2025zoo3d}, we align the reconstructed geometry to the benchmark coordinate system by matching the first predicted pose to the first ground-truth pose and calibrating the global scale using the first-frame depth.

\subsection{Results and Comparisons}
\label{subsec:results}

We compare Group3D with prior open-vocabulary 3D detection approaches under two regimes, \emph{pose-known} and \emph{pose-free}. We focus exclusively on the multi-view RGB setting and report comparisons to both point cloud-based open-vocabulary detectors and multi-view image-based pipelines (Tab.~\ref{tab:scannet_improved}). 

\begin{figure}[tb]
  \centering
  \includegraphics[width=\linewidth]{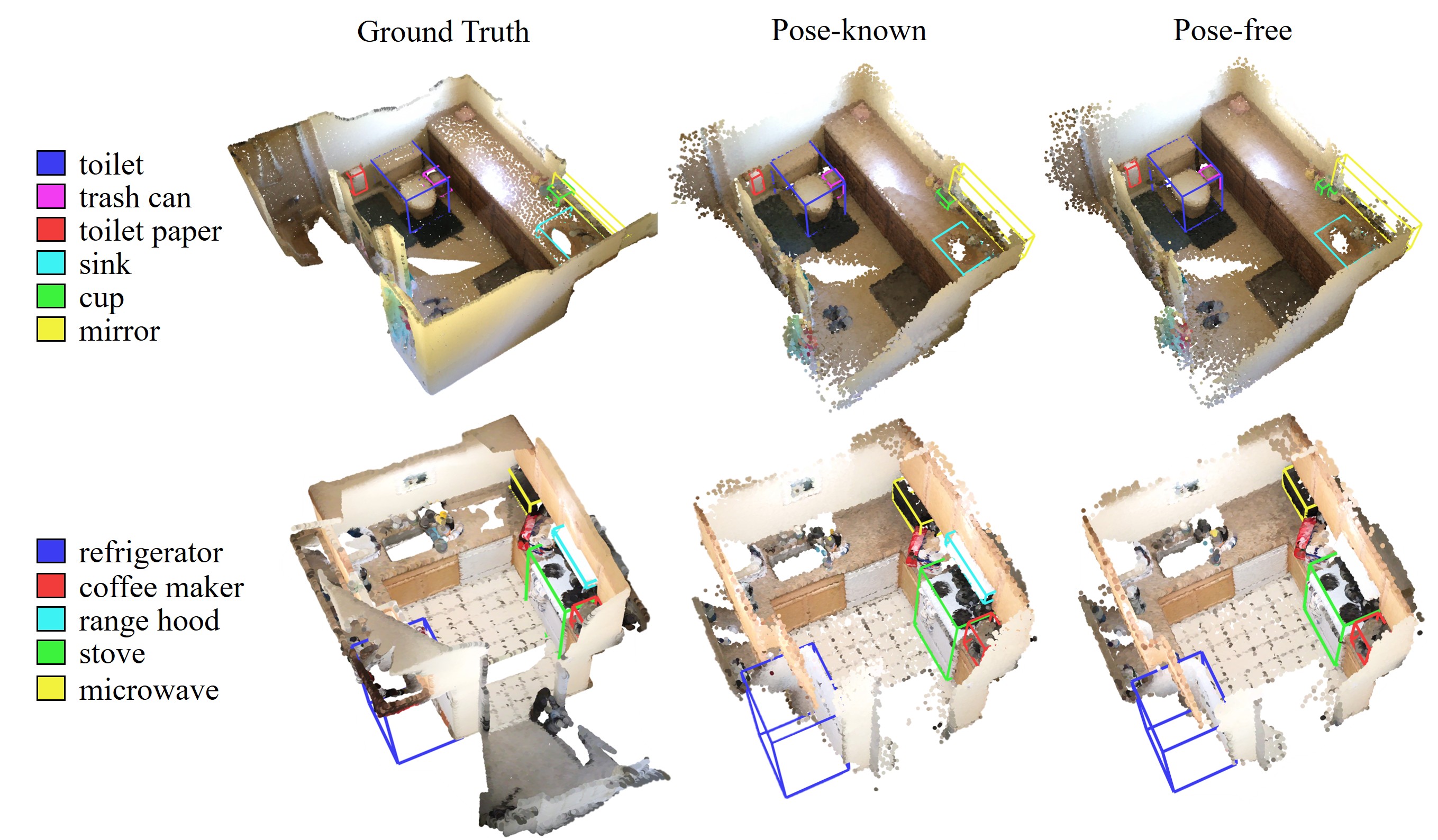}
  \caption{Qualitative results on ScanNet200\cite{rozenberszki2022language} under \emph{pose-known} and \emph{pose-free} settings.}
  \label{fig:qualitative_200}
\end{figure}

On ScanNet20, Group3D establishes a clear new state-of-the-art among multi-view methods. It also surpasses representative approaches that rely on ground-truth point clouds, which highlights that semantically constrained instance construction can compensate for the absence of explicit 3D measurements and even outperform stronger-input baselines. On ScanNet60, Group3D remains competitive and improves over existing multi-view RGB pipelines, suggesting that semantic compatibility grouping continues to provide useful constraints as the vocabulary expands. We additionally evaluate on ScanNet200, a more challenging long-tail setting with a substantially larger and finer-grained vocabulary (Tab.~\ref{tab:scannet200_arkitscenes}). Group3D remains effective under this expanded category space, indicating that MLLM-driven grouping scales beyond a compact taxonomy and supports open-vocabulary recognition in the presence of long-tail categories, as shown in Fig.~\ref{fig:qualitative_200}. In the pose-free regime, where reconstruction noise makes geometry-only association particularly brittle, Group3D preserves strong performance and demonstrates robustness when geometric evidence is incomplete or uncertain. These results highlight the benefit of incorporating semantic compatibility during instance construction, particularly in scenarios where geometric cues alone are insufficient to reliably associate fragments across views. Notably, these improvements are achieved without relying on explicit 3D supervision or ground-truth depth measurements. 
This suggests that combining multi-view geometric cues with language-driven semantic constraints can provide a viable alternative to conventional geometry-driven pipelines.

\begin{figure}[tb]
  \centering
  \includegraphics[width=\linewidth]{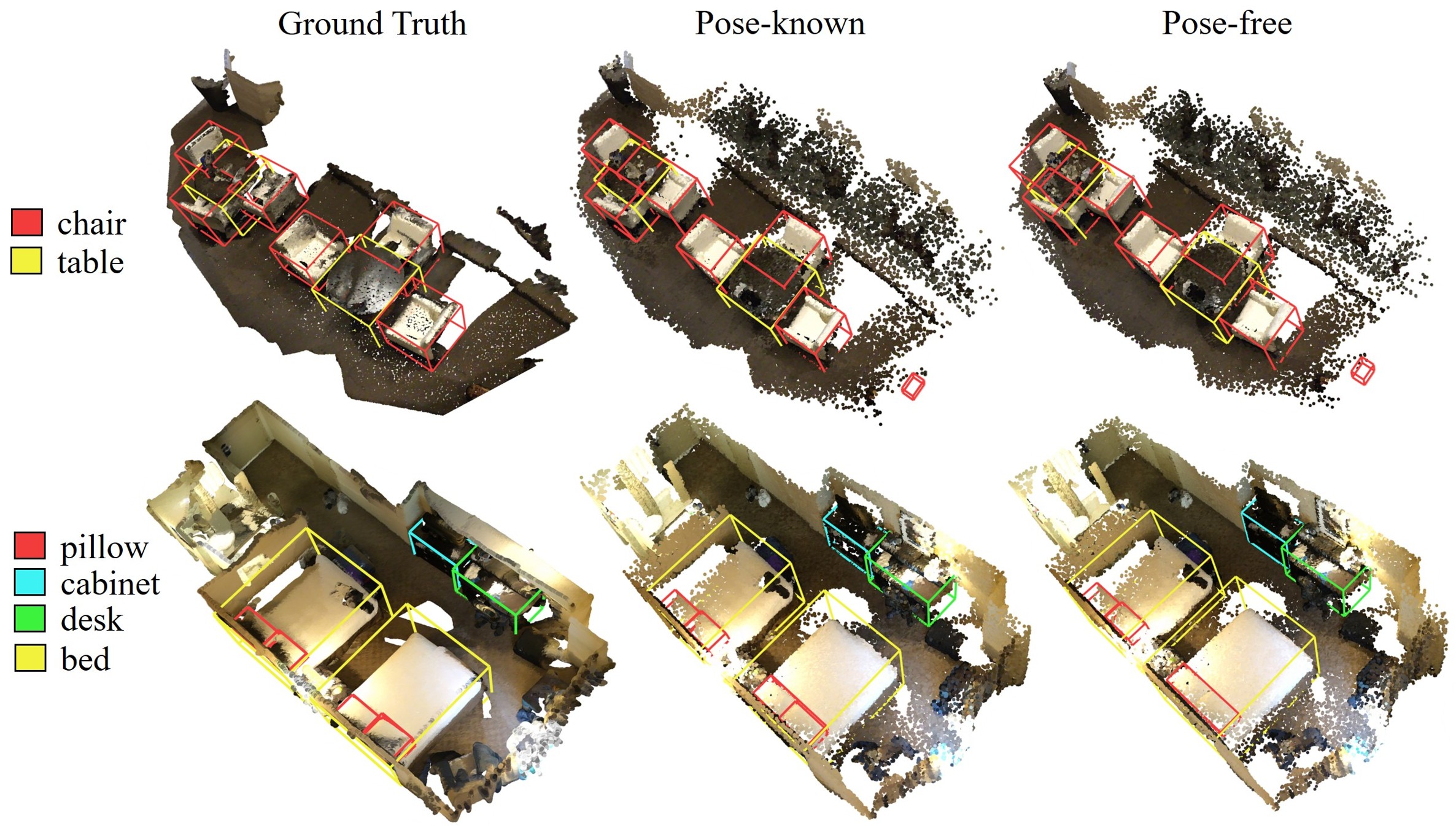}
  \caption{Qualitative results on ScanNet20\cite{lu2023open} under \emph{pose-known} and \emph{pose-free} settings.}
  \label{fig:qualitative_20}
\end{figure}

Table~\ref{tab:ap25} reports per-class results on ScanNet20 and shows that Group3D achieves consistent improvements across a broad set of categories, while Fig.~\ref{fig:qualitative_20} provides visualizations of the predicted open-vocabulary 3D detections. Results on ARKitScenes (Tab.~\ref{tab:scannet200_arkitscenes}) further demonstrate that Group3D generalizes to a different dataset with distinct capture conditions and scene statistics, suggesting that the semantic compatibility prior transfers across domains and supports open-vocabulary 3D detection in the wild.

\begin{table}[h]
\centering
\scriptsize
\caption{
Ablation study on ScanNet20 with different components.
Depth Anything 3 is trained on external datasets, while VGGT is pretrained on ScanNet.
}
\label{tab:ablation}
{
\setlength{\tabcolsep}{8pt}
\renewcommand{\arraystretch}{1.15}
\begin{tabular}{llcc}
\toprule
\textbf{Component} & \textbf{Method} & \textbf{mAP$_{25}$} & \textbf{mAP$_{50}$} \\
\midrule

\multirow{2}{*}{Reconstruction} 
& Depth Anything 3\cite{lin2025depth} & 41.2 & 18.5 \\
& VGGT\cite{wang2025vggt} & 40.0 & 18.7 \\

\midrule

\multirow{2}{*}{MLLM} 
& GPT 5.1\cite{singh2025openai} & 41.2 & 18.5 \\
& Qwen3-VL-8B\cite{bai2025qwen3} & 38.5 & 16.9 \\

\midrule

\multirow{2}{*}{Segmentation} 
& SAM 3\cite{carion2025sam} & 41.2 & 18.5 \\
& Grounded SAM 2\cite{ren2024grounded} & 39.7 & 17.6 \\

\bottomrule
\end{tabular}
}
\end{table}

\begin{table}[h]
\centering
\scriptsize
\begin{minipage}[t]{0.49\linewidth}
\centering
\caption{Ablation on ScanNet20 with different grouping strategies.}
\label{tab:ablation_grouping}
\begin{tabular}{lcc}
\toprule
\textbf{Grouping Strategy} & \textbf{mAP$_{25}$} & \textbf{mAP$_{50}$} \\
\midrule
w/o Category & 28.2 & 9.9 \\
\midrule
Same Category & 35.9 & 14.8 \\
Semantic Compatibility Group & 41.2 & 18.5 \\
\bottomrule
\end{tabular}
\end{minipage}
\hfill
\begin{minipage}[t]{0.49\linewidth}
\centering
\caption{Ablation on ScanNet20 with different K of object category hypotheses.}
\label{tab:ablation_topk}
\begin{tabular}{lcc}
\toprule
\textbf{K} & \textbf{mAP$_{25}$} & \textbf{mAP$_{50}$} \\
\midrule
$K=5$  & 41.2 & 18.5 \\
$K=10$ & 41.2 & 18.8 \\
\bottomrule
\end{tabular}
\end{minipage}

\vspace{-2mm}
\end{table}
\subsection{Ablation Study}

We analyze the key components of Group3D on ScanNet20. As shown in Tab.~\ref{tab:ablation_topk}, varying the number of category hypotheses per view has limited impact on performance: using $K=5$ or $K=10$ yields comparable results. We therefore adopt $K=5$ for improved efficiency without sacrificing accuracy.

Tab.~\ref{tab:ablation} compares different reconstruction backbones, MLLMs, and segmentation models. Replacing the MLLM with a smaller 8B-scale model leads to a moderate performance drop, yet the overall pipeline remains effective, indicating that the proposed semantic grouping mechanism is not tightly coupled to a specific large-scale language model. For reconstruction, VGGT achieves competitive performance but is trained on ScanNet, whereas Depth Anything~3 maintains strong results in a strictly zero-shot setting. Replacing SAM~3 with Grounded SAM~2 slightly reduces performance due to differences in grounding and confidence formulation, while preserving the overall trend.

Finally, Tab.~\ref{tab:ablation_grouping} underscores the importance of semantic compatibility grouping. Removing category information and merging purely by geometry leads to clear degradation due to geometry-driven over-merging. Enforcing a strict same-category constraint mitigates some errors but remains sensitive to cross-view label variability. In contrast, semantic compatibility grouping achieves the best performance by allowing semantically consistent labels to merge while preventing structurally incompatible associations.

\section{Conclusion}

In this work, we proposed Group3D, a multi-view open-vocabulary 3D object detection framework that incorporates semantic constraints directly into the instance construction process. By organizing scene-adaptive category hypotheses into semantic compatibility groups and enforcing merge-time semantic gating, Group3D mitigates geometry-driven over-merging under incomplete and view-dependent multi-view evidence while remaining robust to cross-view category variability. Overall, the results suggest that injecting semantic compatibility into fragment merging leads to more reliable open-vocabulary 3D instance construction using only multi-view RGB inputs. 

More broadly, our findings suggest that integrating language-driven semantic priors into the instance construction process can complement geometric reasoning in multi-view 3D perception. 
Such integration may provide a scalable pathway toward open-world 3D scene understanding without relying on dense 3D supervision or explicit geometry sensors. We hope that this perspective encourages further exploration of language-guided 3D perception systems that bridge visual observations and semantic reasoning. Future work may explore extending the framework to support richer language descriptions and more complex scene-level reasoning across objects, further strengthening the integration between language understanding and multi-view 3D perception. 

%
%
\bibliographystyle{splncs04}
\bibliography{main}

\begin{thebibliography}{10}
\providecommand{\url}[1]{\texttt{#1}}
\providecommand{\urlprefix}{URL }
\providecommand{\doi}[1]{https://doi.org/#1}

\bibitem{bai2025qwen3}
Bai, S., Cai, Y., Chen, R., Chen, K., Chen, X., Cheng, Z., Deng, L., Ding, W., Gao, C., Ge, C., et~al.: Qwen3-vl technical report. arXiv preprint arXiv:2511.21631  (2025)

\bibitem{baruch2021arkitscenes}
Baruch, G., Chen, Z., Dehghan, A., Dimry, T., Feigin, Y., Fu, P., Gebauer, T., Joffe, B., Kurz, D., Schwartz, A., et~al.: Arkitscenes: A diverse real-world dataset for 3d indoor scene understanding using mobile rgb-d data. arXiv preprint arXiv:2111.08897  (2021)

\bibitem{cao2023coda}
Cao, Y., Yihan, Z., Xu, H., Xu, D.: Coda: Collaborative novel box discovery and cross-modal alignment for open-vocabulary 3d object detection. Advances in Neural Information Processing Systems  \textbf{36},  71862--71873 (2023)

\bibitem{carion2025sam}
Carion, N., Gustafson, L., Hu, Y.T., Debnath, S., Hu, R., Suris, D., Ryali, C., Alwala, K.V., Khedr, H., Huang, A., et~al.: Sam 3: Segment anything with concepts. arXiv preprint arXiv:2511.16719  (2025)

\bibitem{carion2020end}
Carion, N., Massa, F., Synnaeve, G., Usunier, N., Kirillov, A., Zagoruyko, S.: End-to-end object detection with transformers. In: European conference on computer vision. pp. 213--229. Springer (2020)

\bibitem{chen2023voxelnext}
Chen, Y., Liu, J., Zhang, X., Qi, X., Jia, J.: Voxelnext: Fully sparse voxelnet for 3d object detection and tracking. In: Proceedings of the IEEE/CVF conference on computer vision and pattern recognition. pp. 21674--21683 (2023)

\bibitem{cheng2021back}
Cheng, B., Sheng, L., Shi, S., Yang, M., Xu, D.: Back-tracing representative points for voting-based 3d object detection in point clouds. In: Proceedings of the IEEE/CVF conference on computer vision and pattern recognition. pp. 8963--8972 (2021)

\bibitem{dai2017scannet}
Dai, A., Chang, A.X., Savva, M., Halber, M., Funkhouser, T., Nie{\ss}ner, M.: Scannet: Richly-annotated 3d reconstructions of indoor scenes. In: Proceedings of the IEEE conference on computer vision and pattern recognition. pp. 5828--5839 (2017)

\bibitem{deng2021voxel}
Deng, J., Shi, S., Li, P., Zhou, W., Zhang, Y., Li, H.: Voxel r-cnn: Towards high performance voxel-based 3d object detection. In: Proceedings of the AAAI conference on artificial intelligence. vol.~35, pp. 1201--1209 (2021)

\bibitem{gwak2020generative}
Gwak, J., Choy, C., Savarese, S.: Generative sparse detection networks for 3d single-shot object detection. In: European conference on computer vision. pp. 297--313. Springer (2020)

\bibitem{hsu2025openm3d}
Hsu, P.H., Zhang, K., Wang, F.E., Tu, T., Li, M.F., Liu, Y.L., Chen, A.Y., Sun, M., Kuo, C.H.: Openm3d: Open vocabulary multi-view indoor 3d object detection without human annotations. In: Proceedings of the IEEE/CVF International Conference on Computer Vision. pp. 8688--8698 (2025)

\bibitem{huang2025nerf}
Huang, C., Hou, Y., Ye, W., Huang, D., Huang, X., Lin, B., Cai, D.: Nerf-det++: Incorporating semantic cues and perspective-aware depth supervision for indoor multi-view 3d detection. IEEE Transactions on Image Processing  (2025)

\bibitem{huang2021bevdet}
Huang, J., Huang, G., Zhu, Z., Ye, Y., Du, D.: Bevdet: High-performance multi-camera 3d object detection in bird-eye-view. arXiv preprint arXiv:2112.11790  (2021)

\bibitem{jia2021scaling}
Jia, C., Yang, Y., Xia, Y., Chen, Y.T., Parekh, Z., Pham, H., Le, Q., Sung, Y.H., Li, Z., Duerig, T.: Scaling up visual and vision-language representation learning with noisy text supervision. In: International conference on machine learning. pp. 4904--4916. PMLR (2021)

\bibitem{jiao2024unlocking}
Jiao, P., Zhao, N., Chen, J., Jiang, Y.G.: Unlocking textual and visual wisdom: Open-vocabulary 3d object detection enhanced by comprehensive guidance from text and image. In: European Conference on Computer Vision. pp. 376--392. Springer (2024)

\bibitem{lang2019pointpillars}
Lang, A.H., Vora, S., Caesar, H., Zhou, L., Yang, J., Beijbom, O.: Pointpillars: Fast encoders for object detection from point clouds. In: Proceedings of the IEEE/CVF conference on computer vision and pattern recognition. pp. 12697--12705 (2019)

\bibitem{lemeshko2025zoo3d}
Lemeshko, A., Gabdullin, B., Drozdov, N., Konushin, A., Rukhovich, D., Kolodiazhnyi, M.: Zoo3d: Zero-shot 3d object detection at scene level. arXiv preprint arXiv:2511.20253  (2025)

\bibitem{li2023dfa3d}
Li, H., Zhang, H., Zeng, Z., Liu, S., Li, F., Ren, T., Zhang, L.: Dfa3d: 3d deformable attention for 2d-to-3d feature lifting. In: Proceedings of the IEEE/CVF International Conference on Computer Vision. pp. 6684--6693 (2023)

\bibitem{li2023bevdepth}
Li, Y., Ge, Z., Yu, G., Yang, J., Wang, Z., Shi, Y., Sun, J., Li, Z.: Bevdepth: Acquisition of reliable depth for multi-view 3d object detection. In: Proceedings of the AAAI conference on artificial intelligence. vol.~37, pp. 1477--1485 (2023)

\bibitem{li2025go}
Li, Z., Yu, H., Ding, Y., Qiao, J., Azam, B., Akhtar, N.: Go-n3rdet: Geometry optimized nerf-enhanced 3d object detector. In: Proceedings of the Computer Vision and Pattern Recognition Conference. pp. 27211--27221 (2025)

\bibitem{li2024bevformer}
Li, Z., Wang, W., Li, H., Xie, E., Sima, C., Lu, T., Yu, Q., Dai, J.: Bevformer: learning bird’s-eye-view representation from lidar-camera via spatiotemporal transformers. IEEE Transactions on Pattern Analysis and Machine Intelligence  \textbf{47}(3),  2020--2036 (2024)

\bibitem{lin2025depth}
Lin, H., Chen, S., Liew, J., Chen, D.Y., Li, Z., Shi, G., Feng, J., Kang, B.: Depth anything 3: Recovering the visual space from any views. arXiv preprint arXiv:2511.10647  (2025)

\bibitem{liu2025fshnet}
Liu, S., Cui, M., Li, B., Liang, Q., Hong, T., Huang, K., Shan, Y.: Fshnet: Fully sparse hybrid network for 3d object detection. In: Proceedings of the Computer Vision and Pattern Recognition Conference. pp. 8900--8909 (2025)

\bibitem{liu2022petr}
Liu, Y., Wang, T., Zhang, X., Sun, J.: Petr: Position embedding transformation for multi-view 3d object detection. In: European conference on computer vision. pp. 531--548. Springer (2022)

\bibitem{liu2024seed}
Liu, Z., Hou, J., Ye, X., Wang, T., Wang, J., Bai, X.: Seed: A simple and effective 3d detr in point clouds. In: European Conference on Computer Vision. pp. 110--126. Springer (2024)

\bibitem{lu2023open}
Lu, Y., Xu, C., Wei, X., Xie, X., Tomizuka, M., Keutzer, K., Zhang, S.: Open-vocabulary point-cloud object detection without 3d annotation. In: Proceedings of the IEEE/CVF conference on computer vision and pattern recognition. pp. 1190--1199 (2023)

\bibitem{mao2021voxel}
Mao, J., Xue, Y., Niu, M., Bai, H., Feng, J., Liang, X., Xu, H., Xu, C.: Voxel transformer for 3d object detection. In: Proceedings of the IEEE/CVF international conference on computer vision. pp. 3164--3173 (2021)

\bibitem{murez2020atlas}
Murez, Z., Van~As, T., Bartolozzi, J., Sinha, A., Badrinarayanan, V., Rabinovich, A.: Atlas: End-to-end 3d scene reconstruction from posed images. In: European conference on computer vision. pp. 414--431. Springer (2020)

\bibitem{peng2024global}
Peng, X., Bai, Y., Gao, C., Yang, L., Xia, F., Mu, B., Wang, X., Liu, S.: Global-local collaborative inference with llm for lidar-based open-vocabulary detection. In: European Conference on Computer Vision. pp. 367--384. Springer (2024)

\bibitem{qi2019deep}
Qi, C.R., Litany, O., He, K., Guibas, L.J.: Deep hough voting for 3d object detection in point clouds. In: proceedings of the IEEE/CVF International Conference on Computer Vision. pp. 9277--9286 (2019)

\bibitem{qi2017pointnet}
Qi, C.R., Su, H., Mo, K., Guibas, L.J.: Pointnet: Deep learning on point sets for 3d classification and segmentation. In: Proceedings of the IEEE conference on computer vision and pattern recognition. pp. 652--660 (2017)

\bibitem{qi2017pointnet++}
Qi, C.R., Yi, L., Su, H., Guibas, L.J.: Pointnet++: Deep hierarchical feature learning on point sets in a metric space. Advances in neural information processing systems  \textbf{30} (2017)

\bibitem{radford2021learning}
Radford, A., Kim, J.W., Hallacy, C., Ramesh, A., Goh, G., Agarwal, S., Sastry, G., Askell, A., Mishkin, P., Clark, J., et~al.: Learning transferable visual models from natural language supervision. In: International conference on machine learning. pp. 8748--8763. PmLR (2021)

\bibitem{ren2024grounded}
Ren, T., Liu, S., Zeng, A., Lin, J., Li, K., Cao, H., Chen, J., Huang, X., Chen, Y., Yan, F., et~al.: Grounded sam: Assembling open-world models for diverse visual tasks. arXiv preprint arXiv:2401.14159  (2024)

\bibitem{rozenberszki2022language}
Rozenberszki, D., Litany, O., Dai, A.: Language-grounded indoor 3d semantic segmentation in the wild. In: European conference on computer vision. pp. 125--141. Springer (2022)

\bibitem{rukhovich2022fcaf3d}
Rukhovich, D., Vorontsova, A., Konushin, A.: Fcaf3d: Fully convolutional anchor-free 3d object detection. In: European Conference on Computer Vision. pp. 477--493. Springer (2022)

\bibitem{rukhovich2022imvoxelnet}
Rukhovich, D., Vorontsova, A., Konushin, A.: Imvoxelnet: Image to voxels projection for monocular and multi-view general-purpose 3d object detection. In: Proceedings of the IEEE/CVF winter conference on applications of computer vision. pp. 2397--2406 (2022)

\bibitem{rukhovich2023tr3d}
Rukhovich, D., Vorontsova, A., Konushin, A.: Tr3d: Towards real-time indoor 3d object detection. In: 2023 IEEE International Conference on Image Processing (ICIP). pp. 281--285. IEEE (2023)

\bibitem{shi2019pointrcnn}
Shi, S., Wang, X., Li, H.: Pointrcnn: 3d object proposal generation and detection from point cloud. In: Proceedings of the IEEE/CVF conference on computer vision and pattern recognition. pp. 770--779 (2019)

\bibitem{shi2020point}
Shi, W., Rajkumar, R.: Point-gnn: Graph neural network for 3d object detection in a point cloud. In: Proceedings of the IEEE/CVF conference on computer vision and pattern recognition. pp. 1711--1719 (2020)

\bibitem{singh2025openai}
Singh, A., Fry, A., Perelman, A., Tart, A., Ganesh, A., El-Kishky, A., McLaughlin, A., Low, A., Ostrow, A., Ananthram, A., et~al.: Openai gpt-5 system card. arXiv preprint arXiv:2601.03267  (2025)

\bibitem{tseng2023crossdtr}
Tseng, C.Y., Chen, Y.R., Lee, H.Y., Wu, T.H., Chen, W.C., Hsu, W.H.: Crossdtr: Cross-view and depth-guided transformers for 3d object detection. In: 2023 IEEE International Conference on Robotics and Automation (ICRA). pp. 4850--4857. IEEE (2023)

\bibitem{tu2023imgeonet}
Tu, T., Chuang, S.P., Liu, Y.L., Sun, C., Zhang, K., Roy, D., Kuo, C.H., Sun, M.: Imgeonet: Image-induced geometry-aware voxel representation for multi-view 3d object detection. In: Proceedings of the IEEE/CVF International Conference on Computer Vision. pp. 6996--7007 (2023)

\bibitem{wang2022rbgnet}
Wang, H., Shi, S., Yang, Z., Fang, R., Qian, Q., Li, H., Schiele, B., Wang, L.: Rbgnet: Ray-based grouping for 3d object detection. In: Proceedings of the IEEE/CVF conference on computer vision and pattern recognition. pp. 1110--1119 (2022)

\bibitem{wang2025vggt}
Wang, J., Chen, M., Karaev, N., Vedaldi, A., Rupprecht, C., Novotny, D.: Vggt: Visual geometry grounded transformer. In: Proceedings of the Computer Vision and Pattern Recognition Conference. pp. 5294--5306 (2025)

\bibitem{wang2022detr3d}
Wang, Y., Guizilini, V.C., Zhang, T., Wang, Y., Zhao, H., Solomon, J.: Detr3d: 3d object detection from multi-view images via 3d-to-2d queries. In: Conference on robot learning. pp. 180--191. PMLR (2022)

\bibitem{wang2024ov}
Wang, Z., Li, Y., Liu, T., Zhao, H., Wang, S.: Ov-uni3detr: Towards unified open-vocabulary 3d object detection via cycle-modality propagation. In: European Conference on Computer Vision. pp. 73--89. Springer (2024)

\bibitem{wu2024panorecon}
Wu, D., Yan, Z., Zha, H.: Panorecon: Real-time panoptic 3d reconstruction from monocular video. In: Proceedings of the IEEE/CVF Conference on Computer Vision and Pattern Recognition. pp. 21507--21518 (2024)

\bibitem{xie2022planarrecon}
Xie, Y., Gadelha, M., Yang, F., Zhou, X., Jiang, H.: Planarrecon: Real-time 3d plane detection and reconstruction from posed monocular videos. In: Proceedings of the IEEE/CVF Conference on Computer Vision and Pattern Recognition. pp. 6219--6228 (2022)

\bibitem{xu2023nerf}
Xu, C., Wu, B., Hou, J., Tsai, S., Li, R., Wang, J., Zhan, W., He, Z., Vajda, P., Keutzer, K., et~al.: Nerf-det: Learning geometry-aware volumetric representation for multi-view 3d object detection. In: Proceedings of the IEEE/CVF international conference on computer vision. pp. 23320--23330 (2023)

\bibitem{xu2024mvsdet}
Xu, Y., Li, C., Lee, G.H.: Mvsdet: Multi-view indoor 3d object detection via efficient plane sweeps. Advances in Neural Information Processing Systems  \textbf{37},  132824--132842 (2024)

\bibitem{yan2018second}
Yan, Y., Mao, Y., Li, B.: Second: Sparsely embedded convolutional detection. Sensors  \textbf{18}(10), ~3337 (2018)

\bibitem{yang2018pixor}
Yang, B., Luo, W., Urtasun, R.: Pixor: Real-time 3d object detection from point clouds. In: Proceedings of the IEEE conference on Computer Vision and Pattern Recognition. pp. 7652--7660 (2018)

\bibitem{yang2024imov3d}
Yang, T., Ju, Y., Yi, L.: Imov3d: Learning open vocabulary point clouds 3d object detection from only 2d images. Advances in Neural Information Processing Systems  \textbf{37},  141261--141291 (2024)

\bibitem{yang2019std}
Yang, Z., Sun, Y., Liu, S., Shen, X., Jia, J.: Std: Sparse-to-dense 3d object detector for point cloud. In: Proceedings of the IEEE/CVF international conference on computer vision. pp. 1951--1960 (2019)

\bibitem{yin2021center}
Yin, T., Zhou, X., Krahenbuhl, P.: Center-based 3d object detection and tracking. In: Proceedings of the IEEE/CVF conference on computer vision and pattern recognition. pp. 11784--11793 (2021)

\bibitem{zhang2025boosting}
Zhang, R., Yu, Z., Cao, S.Y., Zhu, L., Zhang, G., Bai, X., Shen, H.L.: Boosting multi-view indoor 3d object detection via adaptive 3d volume construction. In: Proceedings of the IEEE/CVF International Conference on Computer Vision. pp. 5980--5989 (2025)

\bibitem{zhang2020h3dnet}
Zhang, Z., Sun, B., Yang, H., Huang, Q.: H3dnet: 3d object detection using hybrid geometric primitives. In: European conference on computer vision. pp. 311--329. Springer (2020)

\bibitem{zhou2018voxelnet}
Zhou, Y., Tuzel, O.: Voxelnet: End-to-end learning for point cloud based 3d object detection. In: Proceedings of the IEEE conference on computer vision and pattern recognition. pp. 4490--4499 (2018)

\bibitem{zhu2023pointclip}
Zhu, X., Zhang, R., He, B., Guo, Z., Zeng, Z., Qin, Z., Zhang, S., Gao, P.: Pointclip v2: Prompting clip and gpt for powerful 3d open-world learning. In: Proceedings of the IEEE/CVF international conference on computer vision. pp. 2639--2650 (2023)

\end{thebibliography}


\clearpage
\begin{center}
{\Large \bfseries
Group3D: MLLM-Driven Semantic Grouping for Open-Vocabulary 3D Object Detection\par}
\vspace{0.5em}
{\large Supplementary Material\par}
\end{center}

\appendix

\section{Additional Implementation Details}
\label{sec:intro}

\subsection{Scene Vocabulary Memory}

For each input view, we query the MLLM to obtain a small set of object category hypotheses. The predictions are aggregated across views to form the Scene Vocabulary Memory used throughout the pipeline. The instruction used for this query is shown below.

\begin{tcolorbox}[colback=gray!5,colframe=gray!40,boxrule=0.5pt]

\textbf{Instruction for Scene Vocabulary}

\small
\begin{verbatim}
You are identifying the dominant object categories present
in the scene.

# Task
- Identify the main object categories visible in the image.

# Constraints
- Focus on the most prominent objects in the scene.
- Use simple singular nouns.
- Mention each category only once.
- Avoid descriptive modifiers.

# Output format
- Return a single comma-separated line containing at most
  five object categories.
\end{verbatim}

\end{tcolorbox}

\subsection{Semantic Compatibility Grouping}
We query the MLLM to construct semantic compatibility groups from the scene vocabulary. The list of category names in the scene vocabulary memory is provided to the model together with the instruction below. It encourages grouping categories that may refer to the same physical object across views while avoiding structural or part–whole associations.

\begin{tcolorbox}[colback=gray!5,colframe=gray!40,boxrule=0.5pt]

\textbf{Instruction for Semantic Compatibility Grouping}

\small
\begin{verbatim}
You are generating a semantic merge prior for 3D voxel-based
fragment merging.

# Context
- Categories originate from per-frame 2D class-aware segmentation.
- The same physical object may receive different category names 
  across frames due to taxonomy variations.

# Task
- Group categories that could plausibly refer to the same physical 
  object observed across views.

# Constraints
- Do not group categories merely because they frequently co-occur
  in the same scene or belong to the same structure.
- Do not group structural elements with their openings.
- Do not group part–whole relations.

# Output constraints
- Use only categories from the provided list.
- Each category may appear in at most one group.
- Output only groups containing two or more categories.
- Categories not mentioned are treated as singleton groups.
- Do not include explanations.

# Output format
- group_name: [category1, category2, ...]

\end{verbatim}

\end{tcolorbox}

\subsection{Voxelization and Geometric Overlap}

To measure geometric consistency during fragment merging, each fragment point cloud is discretized into voxels using a fixed voxel size. Given a fragment point cloud $F$, its voxel representation $\texttt{vox}(F)$ is obtained by mapping the 3D coordinates to voxel indices:
\begin{equation}
\texttt{vox}(F)=
\left\{
\left\lfloor \frac{p}{s} \right\rfloor
\;\middle|\;
p \in F
\right\},
\end{equation}
where $p \in \mathbb{R}^3$ denotes a 3D point in the fragment point cloud $F$, and $s$ denotes the voxel size.
In all experiments, we use a voxel size of $5\,\mathrm{cm}$. Geometric overlap between fragments is determined using the overlap predicate defined in Eq.~(6) of the main paper, with thresholds $\tau_{\text{iou}} = 0.01$ and $\tau_{\text{cont}} = 0.10$.

\section{Additional Experiments}

\subsection{More Ablation Studies}
We conduct all ablation studies under the \emph{pose-free} setting, as it represents the more challenging scenario. We further analyze several key components of the proposed pipeline, including the voxel resolution used for fragment merging and the number of input frames.

Tab.~\ref{tab:supp_ablation_voxel} evaluates the effect of voxel size when computing geometric overlap between fragments. Smaller voxels provide more precise alignment and slightly improve detection accuracy. However, the difference between $1\,\mathrm{cm}$ and $5\,\mathrm{cm}$ is marginal, while a larger voxel size (e.g., $10\,\mathrm{cm}$) significantly degrades fragment association due to reduced spatial resolution. Considering the higher computational cost of finer voxelization, we adopt $5\,\mathrm{cm}$ as a practical trade-off between accuracy and efficiency.

Tab.~\ref{tab:supp_ablation_frames} studies the influence of the number of input frames. Increasing the number of views improves scene coverage and produces more complete fragment observations, which benefits instance construction. As the number of frames decreases, the reconstructed geometry becomes less complete and detection performance gradually degrades. Based on this observation, we use $128$ frames in the final configuration.

\begin{table}[h]
\centering
\scriptsize
\begin{minipage}[t]{0.49\linewidth}
\centering
\caption{Ablation on ScanNet20 with different voxel size.}
\label{tab:supp_ablation_voxel}
\begin{tabular}{lcc}
\toprule
\textbf{Voxel Size} & \textbf{mAP$_{25}$} & \textbf{mAP$_{50}$} \\
\midrule
1 cm  & 41.3 & 18.9 \\
5 cm  & 41.2 & 18.5 \\
10 cm & 37.9 & 15.1 \\
\bottomrule
\end{tabular}
\end{minipage}
\hfill
\begin{minipage}[t]{0.49\linewidth}
\centering
\caption{Ablation on ScanNet20 with different number of input frames.}
\label{tab:supp_ablation_frames}
\begin{tabular}{lcc}
\toprule
\textbf{\# Frames} & \textbf{mAP$_{25}$} & \textbf{mAP$_{50}$} \\
\midrule
32 frames & 36.2 & 15.9 \\
64 frames & 39.6 & 17.6 \\
128 frames & 41.2 & 18.5 \\
\bottomrule
\end{tabular}
\end{minipage}

\vspace{-2mm}
\end{table}

\subsection{More Qualitative Results}
\begin{figure}[tb]
  \centering
  \includegraphics[width=\linewidth]{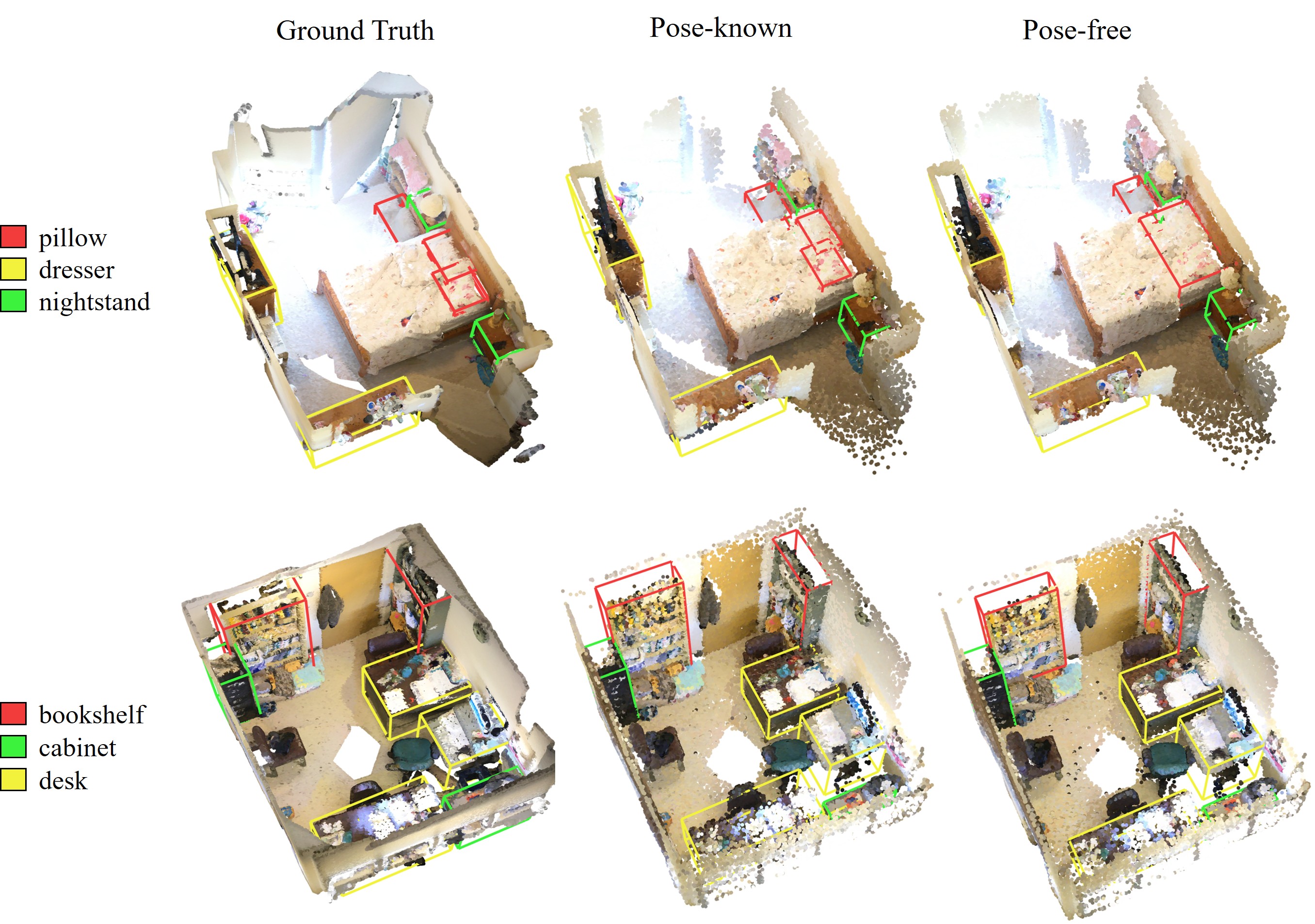}
  \caption{Qualitative results on ScanNet20\cite{lu2023open} under \emph{pose-known} and \emph{pose-free} settings.}
  \label{fig:supp_qualitative_20}
\end{figure}

\begin{figure}[tb]
  \centering
  \includegraphics[width=\linewidth]{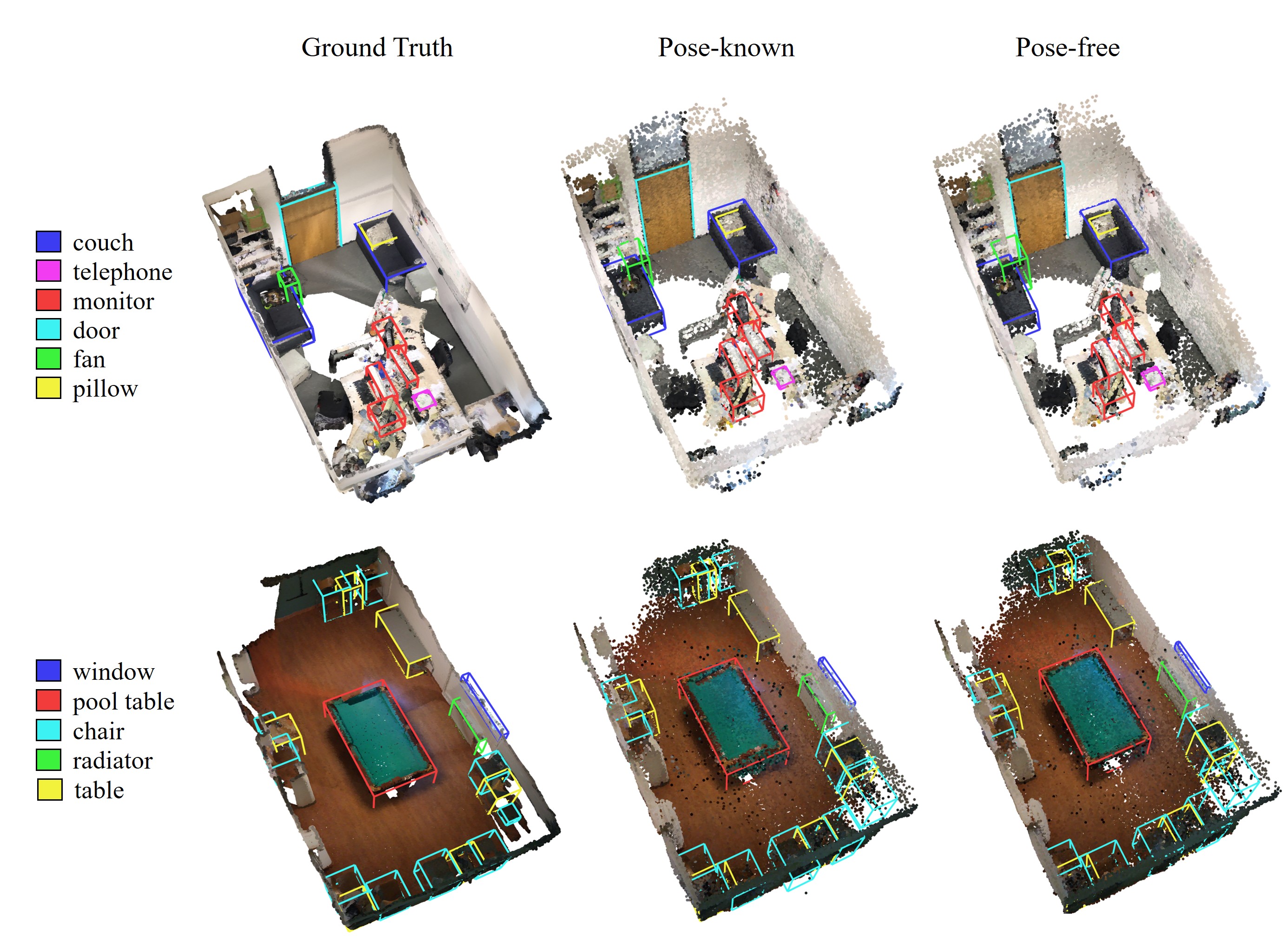}
  \caption{Qualitative results on ScanNet200\cite{rozenberszki2022language} under \emph{pose-known} and \emph{pose-free} settings.}
  \label{fig:supp_qualitative_200}
\end{figure}

\begin{figure}[tb]
  \centering
  \includegraphics[width=\linewidth]{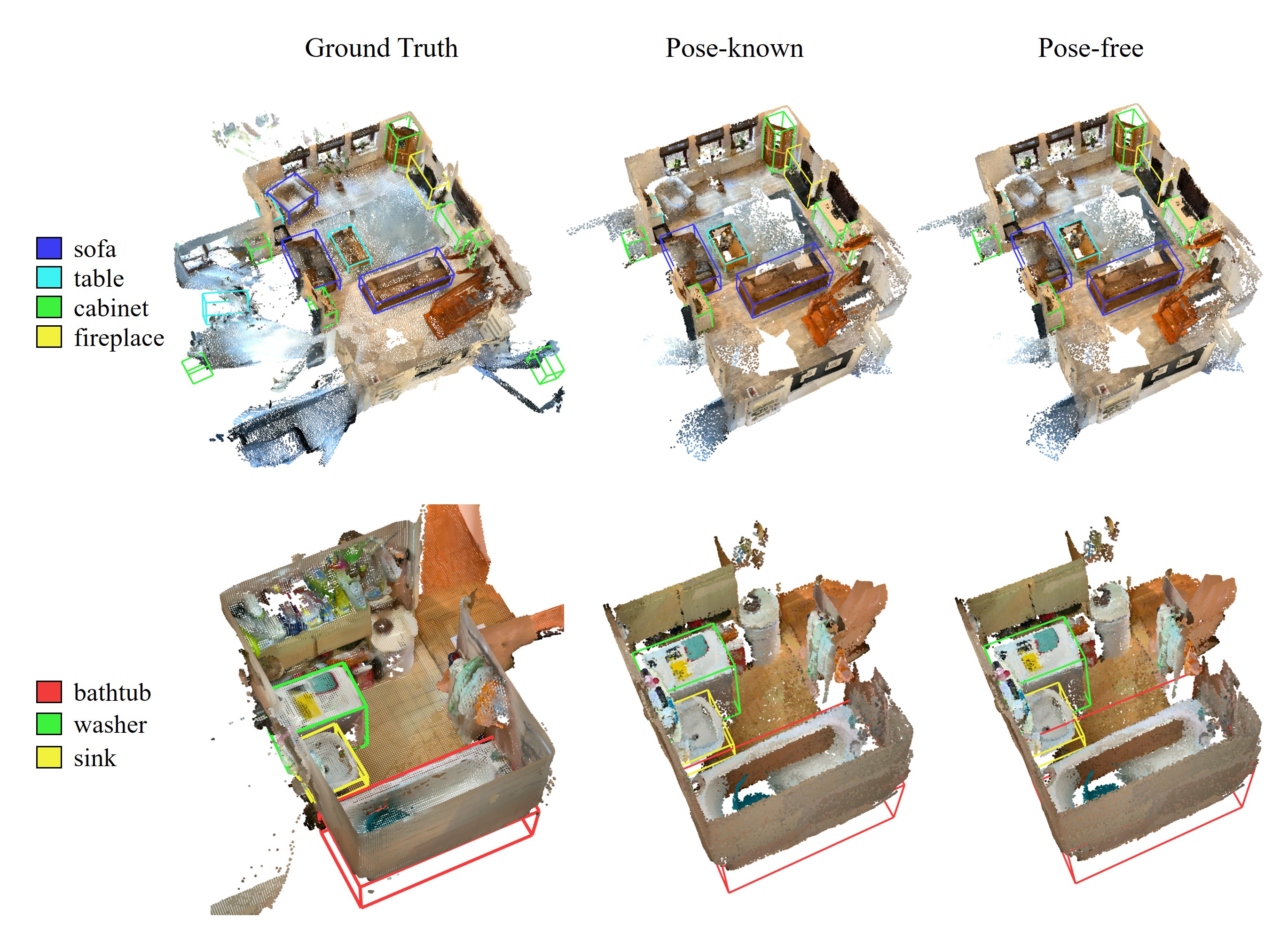}
  \caption{Qualitative results on ARKitScenes\cite{baruch2021arkitscenes} under \emph{pose-known} and \emph{pose-free} settings.}
  \label{fig:supp_qualitative_arkit}
\end{figure}

Additional qualitative results are shown in 
Figs.~\ref{fig:supp_qualitative_20}, \ref{fig:supp_qualitative_200},
and \ref{fig:supp_qualitative_arkit}.
These examples span multiple datasets with diverse vocabulary settings,
including ScanNet20~\cite{lu2023open}, ScanNet200~\cite{rozenberszki2022language}, and ARKitScenes~\cite{baruch2021arkitscenes}.

\subsection{Qualitative Analysis of Semantic Compatibility Grouping}

Tab.~\ref{tab:supp_grouping_examples} presents representative examples of semantic compatibility groupings generated by MLLM from the scene vocabulary memory. The model groups lexical variants and semantically related categories that may refer to the same physical object across views, providing useful semantic priors for the subsequent fragment merging process.

\begin{table}[t]
\centering
\scriptsize
\renewcommand{\arraystretch}{1.1}
\caption{Examples of semantic compatibility grouping predicted by the MLLM
from the scene vocabulary memory.}
\label{tab:supp_grouping_examples}
\begin{tabular}{p{0.5\linewidth}@{\hspace{1.3em}}p{0.5\linewidth}}
\toprule
\textbf{Scene Vocabulary Memory} & \textbf{Semantic Compatibility Grouping} \\

\midrule

bin, book, box, cabinet, clock, copier, door, light switch,
poster, printer, shelf, switch, table, trash can,
washer, washing machine, wall, window
&
\begin{tabular}[t]{@{}l@{}}
\textit{copier\_printer}: [copier, printer] \\
\textit{washer\_machine}: [washer, washing machine] \\
\textit{trash\_container}: [trash can, bin] \\
\textit{switch\_control}: [light switch, switch]
\end{tabular}
\\

\midrule

blackboard, book, bookshelf, box, cabinet, chalkboard,
chair, couch, desk, recliner, shelf, sofa, table, trash can
&
\begin{tabular}[t]{@{}l@{}}
\textit{seating}: [chair, sofa, couch, recliner] \\
\textit{board}: [chalkboard, blackboard] \\
\textit{table\_like}: [table, desk] \\
\textit{storage}: [bookshelf, shelf, cabinet]
\end{tabular}
\\

\midrule

basket, bottle, box, can, carpet, container, cushion, door, jar,
pillow, refrigerator, rug, tray, water bottle, window
&
\begin{tabular}[t]{@{}l@{}}
\textit{small\_containers}: [box, container, basket, tray] \\
\textit{drink\_containers}: [bottle, water bottle, can, jar] \\
\textit{soft\_items}: [pillow, cushion] \\
\textit{ground\_cover}: [carpet, rug] \\
\end{tabular}
\\

\midrule

bag, backpack, bathtub, blanket, computer, lamp, light,
laptop, painting, phone, picture, pillow, shower, sheet,
shoe, sneaker, suitcase, tablet, telephone,
tissue, toilet paper, towel, trash can
&
\begin{tabular}[t]{@{}l@{}}
\textit{computer\_device}: [laptop, computer, tablet] \\
\textit{phone\_device}: [phone, telephone] \\
\textit{bag\_like}: [bag, backpack, suitcase] \\
\textit{footwear}: [shoe, sneaker] \\
\textit{wall\_art}: [painting, picture] \\
\textit{bedding\_cover}: [blanket, sheet] \\
\textit{lamp\_light}: [lamp, light] \\
\textit{bath\_fixture}: [bathtub, shower] \\
\textit{bath\_tissue}: [toilet paper, tissue]
\end{tabular}
\\

\bottomrule
\end{tabular}
\end{table}

\clearpage

\subsection{Open-Vocabulary 3D Instance Segmentation}

Although Group3D is primarily designed for open-vocabulary 3D object detection, the proposed fragment merging process naturally yields instance-level 3D point sets during instance construction. We therefore additionally evaluate the segmentation quality of the reconstructed instances.

Unlike conventional 3D instance segmentation methods that operate directly on the original scene geometry (e.g., point clouds or meshes), our method predicts instances on a newly reconstructed 3D point set obtained from multi-view RGB observations. While this design enables the framework to operate in both pose-known and pose-free settings using RGB inputs alone, it also introduces a geometric mismatch between the reconstructed points and the ground-truth mesh used in ScanNet annotations. Consequently, predicted instance labels must be
transferred to the ground-truth mesh vertices before evaluation.

\paragraph{Evaluation protocol.}
We denote the set of predicted instances produced by Group3D as $\mathcal{C}=\{(C_F, C_{\ell})\}$, where $C_F$ represents the set of 3D points belonging to an instance and $C_{\ell}$ denotes its associated category labels. Following common practice in reconstruction-based 3D scene understanding~\cite{murez2020atlas,xie2022planarrecon,wu2024panorecon}, we assign predicted instance labels to ground-truth mesh vertices using nearest-neighbor association with the reconstructed instance points. For each ground-truth vertex, we find the nearest predicted point among the reconstructed instance points and transfer the corresponding instance label when the nearest-point distance is smaller than $5\,\mathrm{cm}$. To suppress spurious assignments caused by sparse reconstruction or noisy geometry, we additionally require the vertex to lie within the axis-aligned bounding box of the matched instance. Vertices that do not satisfy these conditions are treated as unassigned.

\paragraph{Metric and results.}
After transferring predicted instance labels to ground-truth mesh vertices, we evaluate the resulting vertex-level predictions using the standard ScanNet instance segmentation protocol and report AP$_{25}$ and AP$_{50}$. The results in Tab.~\ref{tab:segmentation} show that Group3D can produce consistent instance-level segmentations despite operating on reconstructed geometry rather than the original scene point cloud. As expected, performance drops in the pose-free setting due to reconstruction noise, but the model still produces valid instance predictions in a zero-shot manner.

\begin{table}[h]
\centering
\scriptsize
\caption{
3D instance segmentation results on ScanNet200~\cite{rozenberszki2022language}.
}
\label{tab:segmentation}
{
\setlength{\tabcolsep}{6pt}
\renewcommand{\arraystretch}{1.05}

\begin{tabular}{l c c cc}
\toprule
\multirow{2}{*}{\textbf{Method}} 
& \multirow{2}{*}{\textbf{Pose-free}}
& \multirow{2}{*}{\textbf{Zero-shot}} 
& \multicolumn{2}{c}{\textbf{ScanNet200}} \\
\cmidrule(lr){4-5}
& & 
& \textbf{AP$_{25}$} & \textbf{AP$_{50}$} \\
\midrule

\multirow{2}{*}{Group3D (Ours)}
& \xmark & \cmark & 22.9 & 12.2 \\
& \cmark & \cmark & 14.8 & 5.3  \\

\bottomrule
\end{tabular}
}
\end{table}



\end{document}